\documentclass[11pt,a4paper]{article}
\usepackage[hyperref]{emnlp2020}
\usepackage{times}
\usepackage{latexsym}

\usepackage{microtype}

\usepackage{adjustbox}
\usepackage{tabularx}
\usepackage{subfigure}
\usepackage{array, multirow}
\usepackage{makecell}
\usepackage[normalem]{ulem}
\usepackage{arydshln}
\usepackage{comment}
\usepackage{hhline}

\aclfinalcopy 
\def\aclpaperid{36} 

\title{Impact of ASR on Alzheimer's Disease Detection:\\All Errors are Equal, but Deletions are More Equal than Others}

\author{Aparna Balagopalan \\
   Winterlight Labs \\
   Toronto, Canada \\
   \texttt{\footnotesize{aparna@winterlightlabs.com}} \\
   \And
   Ksenia Shkaruta \\
   Georgia Tech \\
   Atlanta, USA \\
   \texttt{
   \footnotesize{k.shkaruta@gatech.edu}}\\
   \And
   Jekaterina Novikova \\
   Winterlight Labs \\
   Toronto, Canada \\
   \texttt{\footnotesize{jekaterina@winterlightlabs.com}} \\
   }

\date{}

\begin{document}
\maketitle
\begin{abstract}
Automatic Speech Recognition (ASR) is a critical component of any fully-automated speech-based dementia detection model. However, despite years of speech recognition research, little is known about the impact of ASR accuracy on dementia detection. In this paper, we experiment with controlled amounts of artificially generated ASR errors and investigate their influence on dementia detection. We find that deletion errors affect detection performance the most, due to their impact on the features of syntactic complexity and discourse representation in speech. We show the trend to be generalisable across two different datasets for cognitive impairment detection. As a conclusion, we propose optimising  the  ASR  to  reflect  a higher  penalty  for  deletion  errors  in  order  to improve  dementia  detection  performance.
\end{abstract}

\section{Introduction}
There is a rapid growth in the number of people living with Alzheimer’s disease (AD) 
\cite{alzheimer20182018}. 
Clinical research has shown that quantifiable signs of cognitive decline associated with AD and mild cognitive impairment (MCI) are detectable 
in spontaneous speech \cite{bucks2000analysis,sajjadi2012abnormalities}. 
Machine learning (ML) models have proved to be successful in detecting AD using 
speech and language variables, such as 
syntactic and lexical complexity of language extracted from the transcripts of the speech~\cite{fraser2016linguistic, meilan2012acoustic, rentoumi2014features}. 
Since transcripts 
should  be accurate enough to properly represent  
syntactic and linguistic characteristics, current approaches \cite{fraser2013,zhu2018detecting} frequently rely on 100\% accurate human-created transcripts  produced  by trained transcriptionists. 
However in real-life speech-based applications of AD detection, ASR is used and it produces noisy, error-prone transcripts \cite{yousaf2019comprehensive}.
To our best knowledge, while the importance of well-performing ASR in speech classification has been studied in depth~\cite{zhou2016}, no prior research was done to understand what patterns of 
speech are influenced the most by ASR errors such as word deletions and substitutions, and how this impacts performance of AD detection using ML models.

In this paper, we focus on this issue and study the \emph{effect of deletion, insertion and substitution errors on lexico-syntactic language features} 
and their resulting \emph{effect on classification performance}. 
The effect of these errors on binary AD-healthy classification performance is studied and suggestions are provided on how to improve ASR in order to maintain reasonable AD classification performance.

We identify that deletion errors affect the classification more than substitution and insertion errors on two datasets of spontaneous impaired speech. The effect of these deletion errors are most profound on features related to syntactic complexity and discourse representations in speech, such as production rules, word-level structure and repetitions. These features are also identified as being the most important for the classification task using a feature gradient-based importance metric. 

\section{Data and Setup}
\subsection{Datasets}
\label{sec:datasets}

\textbf{DementiaBank (DB)}
The DementiaBank\footnote{https://dementia.talkbank.org} dataset is a large 
dataset of pathological speech. It consists of narrative picture descriptions from participants aged between 45 to 90 \cite{becker1994natural}. 
Out of the 210 participants in the study, 117 were diagnosed with AD (180 samples of speech) and 93 were healthy (HC, 229 samples). 
Voice recordings and manual transcriptions (following CHAT protocol \cite{macwhinney2000childes}) are available for all samples. This dataset is used for the experiments in Section~\ref{sec:classification_performance}, \ref{sec:deletion_features}, and \ref{sec:model_gradient}.

\textbf{Healthy Aging (HA)}
The Healthy Aging dataset~\cite{balagopalan2018effect} consists of speech samples of 97 participants with no cognitive impairment diagnosis, all older than 50 years. Every participant describes a picture, analogous to the DB dataset. 
The dataset constitutes 8.5 hours of audio with manual transcriptions. Each speech sample is associated with a score on the Montreal Cognitive Assessment (MoCA)~\cite{nasreddine2005montreal}. 
Based on published cut-off scores~\cite{nasreddine2005montreal} for presence of MCI (minimum score for healthy participants is 26), we obtain class-labels for this dataset.


\subsection{ASR Setup}
The  Automatic  Speech  Recognition  (ASR) system we use for this work is based on the open-source  Kaldi  toolkit  \cite{povey2011kaldi}. 
ASR uses ASPiRE chain model trained on multi-condition Fisher English corpus as a 3-gram language model.

\begin{table}[t]
\centering
\scriptsize
\begin{adjustbox}{max width=0.4\textwidth}
\begin{tabular}{ll|l|l|l}
\multicolumn{2}{c|}{\textbf{Dataset}} & \multicolumn{1}{c|}{\textbf{Del (\%)}} & \multicolumn{1}{c|}{\textbf{Ins (\%)}} & \multicolumn{1}{c}{\textbf{Sub (\%)}} \\ \hline \hline
\multirow{2}{*}{DB} & HC & 54.14 & 4.27 & 41.59 \\
 & AD & 56.98 & 3.89 & 39.13 \\ \hdashline[0.5pt/2pt]
\multirow{2}{*}{HA} & HC & 24.37 & 13.11 & 62.52 \\
 & MCI & 21.78 & 14.81 & 63.40 \\ \hline
\end{tabular}
\end{adjustbox}
\caption{\label{tab:transcript_performance} Rates of ASR errors on DB and HA datasets.}
\end{table}

Rates of ASR errors for healthy and impaired speakers for DB and HA datasets are in Table~\ref{tab:transcript_performance}. Majority of errors arise from deletions and substitutions for both datasets and groups.

\section{Methodology}
\label{sec:method}

\subsection{Feature Extraction and Aggregation}
\label{sec:feature_aggregation}
Following previous studies \cite{fraser2016linguistic,balagopalan2018effect}, we automatically extract 507 lexico-syntactic and acoustic features. To simplify the presentation, the extracted features are aggregated into the following major groups:

\hspace{-1em}\textbf{Syntactic Complexity:} features to analyze  the  syntactic  complexity  of speech, such  as  number  of  occurrence  of  various  production  rules,  mean length of clause (in words) etc.

\hspace{-1em}\textbf{Lexical Complexity and Richness :} measures  of lexical  density  and  variation, such as average familiarity scores of all nouns, age of word acquisition, 
frequency of 
POS tags etc.

\hspace{-1em}\textbf{Discourse mapping:} features that help identify cohesion in speech using a \textit{speech graph}-based representation of message organization in speech \cite{mota2012speech}. 
Examples of features include the number of edges  in  the  graph,  number  of  self-loops, cosine-distance across unique utterances etc.


Additionally, we extract features quantifying difficulty in finding the right words (e.g. filled pauses), measures related to description of content in the picture (e.g. number of content units), coherence in speaking at local and global level, and acoustic measures. such as MFCC and Zero Crossing Rate related voice representations (full list in App.\ref{app:feat}).

\subsection{Error and Noise Addition}
\subsubsection{Artificial ASR Errors}
We introduce artificial ASR errors to understand if any specific error type influences the classification performance more than others. In previous research it was shown that lexical and syntactic groups of features extracted from transcripts of speech have different predictive power in dementia classification \cite{novikova2019}. As such, we hypothesize that different ASR error types may influence the features differently and would cause different effects on classification performance. The non-artificial output of ASR combines the errors of deletion, insertion and substitution in some proportion, thus not allowing analysis of the individual effects of each error type separately. This is why we generate each type of errors artificially.

\subsection{Error Addition Method}

We follow a method similar to the one used by ~\citet{fraser2013} to artificially add errors to manual transcripts at predefined 20\%, 40\% and 60\% WER rates. All altered words \textit{w}, where \textit{w} refers to a word in gold-standard manual transcripts, are selected at random. The following modifications are done: a) \textit{deletion} -  word instance \textit{w} is deleted, b) \textit{insertion} - new word \textit{$w_{1}$} is added after the word \textit{w}, c) \textit{substitution} - word \textit{w} is replaced with a new word \textit{$w_{1}$}. 

For \textbf{\textit{deletion}} we simply delete random words from manual transcript at a specified rate. 

To \textbf{\textit{substitute}} word \textit{w}, we select a unigram from 2,000 most used unigrams from Fisher language model that has the smallest Levenshtein distance with word \textit{w} based on the phonemic model from The Carnegie Mellon Pronouncing on Pronouncing Dictionary 
\cite{weide1998cmu}. If word \textit{w} is not found in the Fisher language model a random unigram from the top 2,000 is used for substitution. 

For \textbf{\textit{insertion}}, we select a word from the bigram list from the language model that has the highest probability to follow after word \textit{w} and insert it if it does not match the following word in transcript. In case of a match, the next most probable word is inserted. If word \textit{w} is not found in bigram list a random unigram is used for insertion.

To verify if simulated errors are a fair approximation of what is seen on a true ASR output, we have calculated the BLEU score~\cite{papineni2002bleu} between the manual and ASR-generated transcripts and compared them to the BLEU score between the manual transcripts and the transcripts with artificially simulated errors. The correlation between these two BLEU scores is strong and significant for both datasets (Spearman $\rho=0.72, p< 0.001$ for DB; $\rho=0.66, p<0.001$ for HA), i.e. transcripts with simulated errors are corrupted with respect to the manual transcripts in a similar manner as the ASR-generated transcripts are.

\subsubsection{Noise Addition}
We perturb all lexico-syntactic features or equivalently features that could be affected by ASR errors such as deletions, insertions, and/or substitutions, to mimic random sources of errors using Gaussian noise. We do this to compare and differentiate from the consequences of ASR errors. This modification is implemented by adding a randomized number to the extracted feature values where the mean of the number added to a given feature is zero and the standard deviation varies depending on the amount of noise we add (see App.\ref{app:noise} for details). 




\subsection{Classification Setup}
\label{sec:model}

\textbf{Model:}
All our experiments are based on predictions obtained from a 2-hidden layer neural network (see App.\ref{app:clf} for details). We chose this model type and parameter-setting since it attained performance on-par with previously published results~\cite{fraser2016linguistic} with 10-fold cross-validation on gold-standard manual DB transcripts. 



\section{Changes in Classification Performance Due to Simulated
Errors}
\label{sec:classification_performance}

We evaluate performance of classifying samples of speech to two classes - AD or healthy - using the DB dataset. 

\begin{figure}[t!]
\centering
    \includegraphics[width=0.8\linewidth]{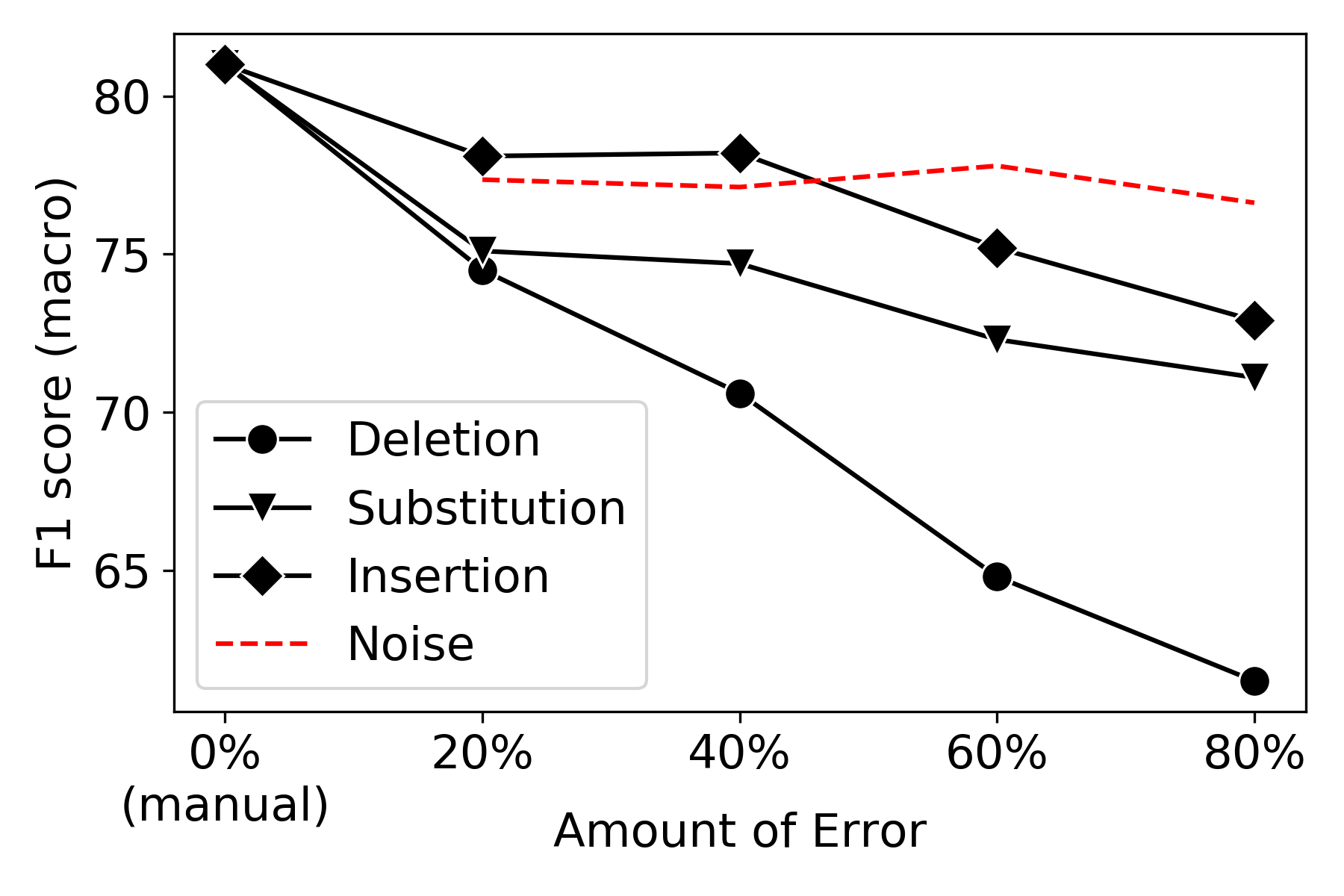}
    \caption{Effect of a controlled amount of ASR errors and random noise on classification performance.}
    \label{fig1:transcript_clf}
\end{figure}

\begin{table}[t]
\setlength\tabcolsep{0pt} 
\footnotesize\centering
\smallskip 
\begin{tabular*}{\columnwidth}{p{0.2\columnwidth}|p{0.2\columnwidth}|p{0.2\columnwidth}|p{0.2\columnwidth}|p{0.2\columnwidth}}
\hline 
      \textbf{Transcript} &  \textbf{Accuracy} & \textbf{F1 (macro)}&\textbf{Sensitivity}&\textbf{Specificity}\\
      \hline \hline
      Manual & 80.20 & 79.76 & 79.29 & 80.83\\
      ASR-based & 74.96 &	73.97 &	76.01 &	74.36\\
      \hline
    \end{tabular*}
 \caption{\label{tab:transcript_asr} Effect of original ASR on classification performance with the DB dataset.}

\end{table}

Figure~\ref{fig1:transcript_clf} shows that deletion errors affect classification performance significantly more than insertion and substitution errors do. 40\% of deletions reduce F1 score by more than 10\%, while 40\% of insertions only result in 2.8\%, and 40\% of substitutions - in 6.3\% of F1 score reduction. These differences become even more pronounced with adding a bigger amount of errors. 
Trajectory of F1 score with varying levels of noise is substantially different from that with varying deletion errors but not that with insertions or substitutions, showing that insertion and substitution errors influence classification performance in a way that is similar to a random noise. Deletion errors, however, have a significantly stronger effect on classification.  It is also interesting to note that the model utilizing automatic transcripts from ASR retains a level of performance at 74.96\% (Table~\ref{tab:transcript_asr}), which is comparable to the potential decrease in performance due to the rate of ASR deletion errors.

Different effects of errors on classification performance suggest that some features, extracted from the speech samples and used as an input for the classification algorithm, are affected far more substantially by deletions rather than any other type of errors. This leads us to inspect the correlation of feature values and the amount of deletions.
\vspace{-0.5em}

\section{Distinctive Effects of Deletion Errors}
\label{sec:deletion_features}

In order to understand why deletions errors influence the classification performance significantly more than other error types, we identify features maintaining higher correlation with the amount of deletions than that with the amount of insertions and substitutions. 
We observe 18 features in total that distinctively correlate with deletions. Out of these, the absolute majority of 15 features (83.33\% of all selected) are associated with syntactic complexity (production rules of a constituency parser) and discourse phenomena (graph self-loop with 3 edges) and 3 (16.7\%) - with lexical richness in speech. Other feature groups, such as acoustic features or those associated with word finding difficulty, do not meet the required conditions.
Such results show that syntactic structure of language is much more vulnerable to deletions than to other ASR errors. This can be explained by the fact that insertions and substitutions use words from the language model (i.e. most probable words) for the modifications, which to some extent helps maintain basic syntactic rules and structure. 

Correlation between the number of deletions and features of syntactic structure shows the vulnerability of the feature group representing syntactic complexity and discourse phenomena to ASR deletion errors. However, it does not explain a decrease in classification performance when adding deletion errors. In Section~\ref{sec:model_gradient} we inspect if features of syntactic complexity are more influential in AD detection than other characteristics of speech.
\vspace{-0.5em}

\section{Model-based Analysis of Feature Importance}
\label{sec:model_gradient}

In order to quantify the importance of input features for classification, we obtain the gradient of the output prediction loss with respect to input features on a manually-transcribed version of the DB dataset.

We define gradient-based importance for feature $k$ for an input, $X_{i,j}$, in the training set for a classification model as:
\begin{equation}
    imp^{i,j,k} = \frac{\partial L(y_{i,j}, p_{i, j})}{\partial X_{i,j,k}}
\end{equation}
\vspace{-0.6em}

where $L$ denotes the loss criterion (binary cross-entropy loss), $y_{i,j}$ is the ground-truth label, $p_{i, j} \subset  [0,1]$ is the prediction probability; $p_{i, j} > 0.5$ denotes an AD prediction, $k$ is a given feature (1 to $D$ ), and $i$ is a number of samples (1 to $N_{j}$) in the training set in fold $j$ of the DB dataset classification setup. 
Hence, to obtain the average importance for feature $k$ in a single fold, we compute:
\begin{equation}
    imp^{j, k} = 1/N_{j}\sum_{i=1}^{N_{j}}\frac{\partial L(y_{i,j}, p_{i, j})}{\partial X_{i,j,k}}
\end{equation}
This importance is then averaged across the 10-folds to obtain the final importance, i.e.:
\begin{equation}
    imp^{k} = 1/10\sum_{j=1}^{10}imp^{j, k}
\end{equation}

In order to interpret high-level patterns of input importance, we aggregate the feature importances into the groups defined in Section~\ref{sec:feature_aggregation}, where aggregation of importances involves averaging the absolute gradient-importance, $|imp^{k}|$, of features belonging to that group.

Results provided in Table~\ref{tab:top10_imp} show that the average normalised importance of the features associated with syntactic complexity and discourse is higher than the average importance of lexical richness features, when top-10 most important features across all the groups are selected for comparison.

\begin{table}[t]
\scriptsize 
\begin{tabular}{l|cc|l|l}
\hline
\multirow{2}{*}{\textbf{Feature group}} & \multicolumn{2}{l}{\textbf{\begin{tabular}[c]{@{}l@{}}Importance of\\ top-10 features\end{tabular}}} & \multirow{2}{*}{\textbf{\#features}} & \multirow{2}{*}{\textbf{Group rank}} \\
 & \textbf{HC} & \textbf{AD} &  &  \\ \hline\hline
\begin{tabular}[c]{@{}l@{}}Syntactic complexity and\\ Discourse phenomena\end{tabular} & 0.94 & 0.95 & \multicolumn{1}{c|}{37} & \multicolumn{1}{c}{1} \\
Lexical richness & 0.91 & 0.92 & \multicolumn{1}{c|}{18} & \multicolumn{1}{c}{2} \\ \hline
\end{tabular}
\caption{\label{tab:top10_imp} Importance of the two feature groups, summarised as the mean value of the top-10 most important features selected for HC and AD components, number of features having significant Spearman correlation with deletion errors, and the rank of each group.}
\end{table}



To conclude, the feature group of syntactic complexity and discourse phenomena is affected significantly and distinctively the most by deletion errors as seen in Section~\ref{sec:deletion_features}. This group is also important for classification as seen in Table~\ref{tab:top10_imp}, indicating why classification is affected significantly by deletion errors. Hence, we track the effects from the initial step of adding artificial errors of different amounts to obtaining the final predictions in this manner.

\vspace{-0.5em}

\section{Generalisability Evaluation}
\label{sec:general}

In order to test how well our conclusions generalise to a different dataset of impaired speech, we repeat the same experiments performed on DB on 
the HA dataset (Section~\ref{sec:datasets}). 

We follow the same method, as described in Section~\ref{sec:method} 
to extract the features and classify samples. 
Similarly to the results obtained on DB data, with HA deletion errors affect classification performance the most. 
Furthermore, deletion errors differentiate the same feature group of syntactic complexity and discourse phenomena
: with HA dataset, 39 features correlate with deletions stronger than with insertions or substitutions, with $79.49$\% of features belonging to the aggregate group of syntactic complexity and discourse, and $20.51$\% - to the group of lexical richness. 
The rank of feature groups, based on the average absolute Spearman correlation of all the features included in the groups, correspond to the rank observed with DB dataset, with a stronger significant correlation corresponding to the group of syntactic complexity, rather than lexical richness. 
\vspace{-0.5em}

\section{Conclusions}
We observe that simulated deletion errors have a strong effect on classification performance when detecting cognitive impairment from speech and language, which can be traced back to their effect on syntactic complexity and discourse representations.  
With this observation in mind, the practical suggestion would be to 
optimise the ASR to reflect a higher penalty for deletion errors to improve dementia detection performance. For example, the decoder can be parametrised to find a balance between insertions and deletions, so that the number of deletion errors is minimised. 

However, dealing with deletions in training time is not trivial, so in future work, we will focus on the optimisation of ASR performance and its effect on AD detection. Careful ASR error management, following previous work by~\citet{simonnet2017asr}, could help enable strong fully-automated speech-based predictive models for dementia detection.



\bibliographystyle{acl_natbib}
\bibliography{icassp2020}

\vfill\pagebreak
\clearpage

\appendix
\setcounter{table}{0}
\renewcommand{\thetable}{A.\arabic{table}}

\section{Appendices}
\label{app}

\subsection{List of Features Extracted from Speech and Transcripts}
\label{app:feat}

Full list of lexico-syntactic, discourse mapping and additional language  features is in Table~\ref{tab:features}, all with brief descriptions and counts of sub-types. Spacy\footnote{https://spacy.io/usage/linguistic-features} is used for part-of-speech tagging in the linguistic pipeline, where tags belong to Penn Treebank\footnote{http://www.cis.upenn.edu/~treebank/}.

\begin{table*}
\centering

{
\begin{adjustbox}{max width=\linewidth}

\begin{tabular}{l|c|l}
 \textbf{Feature type} & \textbf{\#Features} & \textbf{Brief Description} \\
\hhline{===}
Syntactic Complexity & 36 & L2 Syntactic Complexity Analyzer~\cite{lu2010automatic} features; max/min utterance length, depth of syntactic parse tree\\
 &104 & Number of times a production type occurs divided by total number of productions \\
 & 13 & Proportion, average length and rate of phrase types\\
\cline{1-3}
\multirow{2}{*}{Lexical Complexity and Richness}
 & \multirow{2}{*}{12} & Average norms across all words, across nouns only and across verbs only for imageability, \\
& & age of acquisition, familiarity and frequency (commonness)\\
 & 6 & Type-token ratios (including moving window); brunet; Honoré’s statistic\\
\multirow{3}{*}{} & \multirow{3}{*}{5} & Proportion of demonstratives (e.g., "this"), function words,
\\ 
& & light verbs and inflected verbs, and propositions (POS tag verb, adjective, adverb,\\ & & conjunction, or preposition)\\
 & 3 & Ratios nouns:(nouns+verbs); nouns:verbs; pronouns:(nouns+pronouns)\\
 & 1 & Average word length\\
 & 18 & Proportions of Spacy univeral POS tags \\
 & 53 & Proportions of Penn Treebank POS tags\\
\cline{1-3}
Discourse Mapping & 15 & Avg/max/min similarity between word2vec~\cite{mikolov2013distributed} representations of utterances (with different dimensions)\\
 & 5 & Fraction of pairs of utterances below a similarity threshold (0.5,0.3,0); avg/min distance\\
 & 13 & Representing words as nodes in a graph and computing density, number of loops, etc.\\
 & 1 & Number of switches in verb tense across utterances divided by total number of utterances\\
 & 15 & Avg/min/max cosine distance between word2vec~\cite{mikolov2013distributed} utterances and picture content units, with varying dimensions of word2vec \\
\cline{1-3}
Additional Features &  2 & Ratios -- number of words: duration of audio; number of syllables: duration of speech, \\
 & 1 & Proportion of words not in the English dictionary\\
 & 9 & Average sentiment valence, arousal and dominance across all words, noun and verbs\\
 &\multirow{2}{*}{7} &  Total and mean duration of pauses;long and short pause counts;\\
& & pause to word ratio; fillers(um,uh); duration of pauses to word durations\\
 & 4 & Avg/min/max/median fundamental frequency of audio\\
 & 2 & Duration of audio and spoken segment of audio \\
 & 4 & Avg/variance/skewness/kurtosis of zero-crossing rate\\

 & 168 & Avg/variance/skewness/kurtosis of 42 Mel-frequency Cepstral Coefficients (MFCC) coefficients\\
  & 10 & Proportion of lemmatized words, relating to the Cookie Theft picture content units to total number of content units\\
\cline{1-3}

\end{tabular}
\end{adjustbox}
}
\caption{Summary of all lexico-syntactic features extracted. The number of features in each subtype is shown in the second column (titled ``\#Features").\label{tab:features}
}
\end{table*}




\subsection{Noise Addition Method}
\label{app:noise}
The strategy followed to add $k$ level of noise ($\epsilon$) to an input $X_{i}$ is as follows:

\begin{equation}
X^{noised^{k}}_{i,j} = X_{i,j} + \epsilon
\end{equation}
\begin{equation}
    \epsilon \sim \mathcal{N}(0, \sigma_{new}^{2})
\end{equation}
\begin{equation}
     \sigma_{new} = k* \sigma
\end{equation}


where $i$ 
is a sample number, $j$ 
is a feature number, and standard deviation of noise added is $k$ times the standard deviation of the original unperturbed feature, $\sigma$. Note that the 
standard deviation ($\sigma$) per feature is calculated over \emph{all} samples. 

\subsection{Classification Setup Details}
\label{app:clf}
\textbf{Hyperparameter Settings}: All 2 hidden layers of our network have 10 units each~\cite{pedregosa2011scikit}. We use the Adam optimizer \cite{kingma2014adam} with an initial learning rate of $0.001$ and $L2$ regularization parameter for network weights set to $0.0001$. The model is trained to a maximum of 200 epochs. 

\textbf{Evaluation:}
For classification experiments on all datasets (DB and HA, varying levels of artificial errors), we perform 10-fold cross-validation stratified at subject level (ensuring that speech samples from the same subject do not in the training and testing sets in each fold). The F1 (macro) - score, averaged across the 10 folds, is chosen as our primary evaluation metric, given the imbalanced nature of our dataset.

\end{document}